# Evaluation Report on Large Language Models Solving High School Mathematics Questions

Zhu Jiawei[1], Chen Wei[2]

(1. School of Environmental Science and Engineering, Guangdong University of Technology, Guangzhou, Guangdong 510006, China; 2. School of Computer Science, Guangdong University of Technology, Guangzhou, Guangdong 510006, China)

**Abstract**：This report aims to evaluate the performance of large language models (LLMs) in solving high school science questions and to explore their potential applications in the educational field. [1] With the rapid development of LLMs in the field of natural language processing, their application in education has attracted widespread attention. This study selected mathematics exam questions from the college entrance examinations (2019-2023) as evaluation data and utilized at least eight LLM APIs to provide answers. A comprehensive assessment was conducted based on metrics such as accuracy, response time, logical reasoning, and creativity. Through an in-depth analysis of the evaluation results, this report reveals the strengths and weaknesses of LLMs in handling high school science questions and discusses their implications for educational practice. The findings indicate that although LLMs perform excellently in certain aspects, there is still room for improvement in logical reasoning and creative problem-solving. This report provides an empirical foundation for further research and application of LLMs in the educational field and offers suggestions for improvement.

**Keyword**：Large Language Models; High School Mathematics Questions; Educational Applications; Performance Evaluation; Natural Language Processing

## 1 Introduction

In the information age of the 21st century, the development of artificial intelligence technology is advancing rapidly. Among these advancements, large language models (LLMs) have emerged as a breakthrough technology in the field of natural language processing (NLP), becoming a significant force in driving the development of language intelligence. LLMs, through deep learning and training on massive datasets, are capable of understanding and generating natural language, enabling various language tasks such as machine translation, text summarization, and question-answering systems.

In recent years, LLMs have demonstrated immense potential in the educational field. They can serve as teaching aids, helping students understand complex concepts, providing personalized learning recommendations, and even automatically generating or grading assignments and exam questions. Additionally, LLMs can analyze students' learning data to provide feedback to teachers,

---

[1] Zhu Jiawei*（May 2005—）, Male. Chen Wei is the supervising teacher

assisting them in optimizing teaching methods and content.

Although LLMs have shown great potential, their application in the educational field is still in its infancy, and their performance and effectiveness require rigorous evaluation and testing. This study aims to assess the ability of LLMs to solve high school science questions, gaining a deeper understanding of their application potential in education. By designing and implementing a series of evaluations, this study will analyze the performance of LLMs in terms of accuracy, response time, logical reasoning, and creativity when solving high school mathematics questions. This will provide valuable reference information for educators and technology developers, promoting the healthy development of LLMs in the educational field.

## 2 Evaluation Data

### 2.1 Data Sources

2.1.1 The primary data source for this study consists of mathematics college entrance examination papers from 2019 to 2023. These papers are compiled by national educational examination authorities and cover the core knowledge points and typical question types of high school science education. Given the differences in core examination content between 2019-2020 and 2021-2023 due to the reform of the college entrance examination in 2021, this study includes papers containing both pre-reform and post-reform data to enhance the generality and representativeness of the sample. Through these papers, we aim to construct a comprehensive and representative evaluation dataset to assess the performance of large language models (LLMs) in solving high school science questions. This report primarily uses the data in Table 1 as an example.

**Table 1 Papers for which data were selected for this report**

| From 2019 to 2020 | From 2021 to 2023 年 |
|---|---|
| New Curriculum Volume 1 | New College Entrance Examination Volume 1 |
| New Curriculum Volume 2 | New College Entrance Examination Volume 2 |

### 2.2 Data Selection

For the selected data, we categorized each question based on its characteristics as follows:

Firstly, according to different question types, they are divided into the following categories:
- a) Multiple-choice questions (including single-choice and multiple-choice questions)
- b) Fill-in-the-blank questions
- c) Comprehensive questions

Next, according to different difficulty levels, they are categorized as follows:

**Table 2 Different question types correspond to different difficulty factors**

| Basic Questions | DF = 0.7 ~ 1.0 |
|---|---|
| Intermediate Questions | DF = 0.5 ~ 0.7 |
| Advanced Questions | DF = 0.3 ~ 0.5 |
| Difficult Questions | DF = 0.0 ~ 0.3 |

It should be noted that the data for the difficulty factor (DF) is derived from the average of national assessment data for each exam paper, as shown in Table 3.

**Table 3 Difficulty coefficient definition table**

| Basic Questions (DF = 0.7 ~ 1.0) |
|---|
| If a question's difficulty factor is above 0.7, it can be considered an easy question. Most students can answer it correctly, which may indicate that the knowledge point is relatively basic, or the question is clearly stated and easy to understand. |
| Intermediate Questions (DF = 0.5 ~ 0.7) |
| Questions with a difficulty factor below 0.3 are considered difficult. Only a very small number of students can answer them correctly, which may indicate that the knowledge points are very complex, requiring a comprehensive and deep understanding and analysis to solve. |
| Advanced Questions (DF = 0.3 ~ 0.5) |
| Questions with a difficulty factor below 0.3 are considered difficult. Only a very small number of students can answer them correctly, which may indicate that the knowledge points are very complex, requiring a comprehensive and deep understanding and analysis to solve. |
| Difficult Questions (DF = 0.0 ~ 0.3) |
| Questions with a difficulty factor below 0.3 are considered difficult. Only a very small number of students can answer them correctly, which may indicate that the knowledge points are very complex, requiring a comprehensive and deep understanding and analysis to solve. |

Finally, in all the data, we meticulously categorized all characteristic question types in a set of exam papers to reduce testing time while ensuring the data's representativeness. We strictly adhered to the following principles: avoid selecting questions with diagrams as much as possible, distribute difficulty levels evenly across question types, and maintain relative uniformity. Table 4 shows the selection of questions at different difficulty levels across various exam papers.

**Table 4 Distribution of topics**

| | 2019（Ⅰ） | 2019（Ⅱ） | 2020（Ⅰ） | 2020（Ⅱ） | 2021（Ⅰ） |
|---|---|---|---|---|---|
| Basic Questions | 1、13、17 | 1、13、17 | 1、13、17 | 1、13、17 | 1、13、17 |
| Intermediate Questions | 4、14、19 | 4、14、19 | 5、14、19 | 4、14、19 | 4、14、18 |
| Advanced Questions | 10、15、19 | 10、15、19 | 10、15、19 | 9、15、19 | 10、15、20 |
| Difficult Questions | 12、16、20 | 12、16、20 | 12、16、20 | 12、16、20 | 8、16、22 |
| | 2021（Ⅱ） | 2020（Ⅰ） | 2020（Ⅱ） | 2020（Ⅰ） | 2020（Ⅱ） |
| Basic Questions | 1、13、17 | 1、13、17 | 1、13、17 | 1、13、17 | 1、13、17 |
| Intermediate Questions | 4、14、18 | 4、14、18 | 5、14、18 | 4、14、18 | 4、14、18 |
| Advanced Questions | 10、15、20 | 10、15、20 | 10、15、20 | 10、15、19 | 10、15、20 |
| *Difficult Questions* | 8、16、22 | 8、16、22 | 8、16、21 | 8、16、21 | 8、16、21 |

### 2.3 Data Processing

In the report, tables can be used to display the structure of JSON files, providing a more intuitive presentation of data organization, as shown in Table 5.

**Table 5 Table Property Sheet**

| Attribute | Description |
|---|---|
| id | Unique identifier of the question |
| year | Year the question belongs to |
| difficulty | Difficulty level of the question |
| type | Type of the question |
| number | Serial number of the question |
| Source | Source of the question |
| Description | Description of the question, including text and image paths |
| An | Standard answer to the question, stored in LaTeX format |

We organized all questions into a JSON file, with each question stored as an object. Each object contains basic information about the question (such as id, year, difficulty, etc.), as well as a detailed description (Description) and the standard answer (An). The question description includes the text and related image paths, while the answer is stored in a string in LaTeX format.

## 3 Evaluation Selection

### 3.1 Model Selection

When evaluating the performance of large language models (LLMs), we carefully selected the following representative models based on domestic and international classification standards to ensure the comprehensiveness and objectivity of the evaluation results.

**Table 6 List of topics**

| CHINESE | Abroad |
|---|---|
| **Qinghua**（GLM-4-Flash） | Google（gemma-2-9b-it） |
| **Baidu**（ERNIE-Speed-128K） | Meta（Meta-Llama-3.1-8B-Instruct） |
| **iFlytek**(Spark Lite) | |
| **Tencent**(HUNYUAN-lite) | |
| **Alibaba Cloud**(Qwen/Qwen2.5-7B-Instruct) | |
| **Zero One**(Yi-34B) | |

## 3.2 Model Description

**Table 7 Model introduction table**

| | |
|---|---|
| GLM-4-Flash | GLM-4-Flash retains the excellent features of previous models, such as smooth dialogue and low deployment threshold, while introducing new features. It uses more diverse training data, more thorough training steps, and more reasonable training strategies, performing well among pre-trained models under 10B. |
| ERNIE-Speed-128K | ERNIE-Speed-128K, launched by Baidu, is a large language model widely used in search, recommendation, and natural language understanding. It is recognized in the industry for its efficient information retrieval capabilities and accurate semantic analysis. |
| Spark Lite | Spark Lite performs well in tasks such as dialogue generation and text summarization. Spark Lite is noted for its flexible model architecture and efficient training mechanism. |
| HUNYUAN-lite | A large language model launched by Tencent, it excels in content generation, capable of producing rich and diverse text content, and has high accuracy in semantic understanding. |
| google/gemma-2-9b-it | Gemma is one of Google's lightweight, state-of-the-art open model series. It is a decoder-only large language model supporting English, offering open weights, pre-trained variants, and instruction-tuned variants. The 9B model is trained on 8 trillion tokens and is suitable for various text generation tasks, including Q&A, summarization, and reasoning. |
| Yi-34B | The open-source Yi-34B model supports a 200K ultra-long context window version, capable of processing ultra-long text inputs of about 400,000 Chinese characters and understanding PDF documents over 1,000 pages. |
| Qwen/Qwen2.5-7B-Instruct | Qwen2.5-7B-Instruct is one of the latest large language model series released by Alibaba Cloud. The 7B model has significantly improved capabilities in fields such as coding and mathematics. |
| Meta-Llama-3.1-8B-Instruct | Meta Llama 3 is a family of large language models developed by Meta, including 8B and 70B parameter scale pre-trained and instruction-tuned variants. The 8B instruction-tuned model is optimized for dialogue scenarios and performs excellently in multiple industry benchmarks. |

3.2.1 Data Management

To facilitate the management and processing of data returned by different model APIs, we took the following steps: First, we handled the input and output of each model separately. For each model, we output two types of processed results: single-solution responses and multi-solution responses. Single-solution responses are used to evaluate the model's accuracy and logical reasoning, while multi-solution responses are used to assess its creativity.

Table 8 Model Properties Table

| Attribute | Description |
| --- | --- |
| id | Unique identifier for the model's response |
| question | Input question for the model |
| answer | Output answer from the model |
| response_time | Model's response time, in seconds |

These results are stored in different JSON files for subsequent data processing. For example:

```
{
  "question": "xxx",
  "answer": "x",
  "response_time": xx.xx,
  "id": x
}
```

3.2.2 Data Classification

For classifying the types of solutions output by the models, we altered the prompt words given to the models to change their responses. For multi-solution problems, the prompt word is:

Table 9 Prompt word classification table

| Single-Solution PROMPT | "Below is a question; please provide multiple solutions for this question, separating each solution with a newline:" |
| --- | --- |
| Multi-Solution PROMPT | "Below is the complete question. If it is a multiple-choice or fill-in-the-blank question, please provide only the answer " |

## 4 Evaluation Metrics and Methods

### 4.1 Accuracy and Its Evaluation Method

For the accuracy of responses from the same model to different questions, we divided it into single-solution accuracy and multi-solution accuracy.

4.1.1 Single-Solution Accuracy

For single-solution responses, we used different methods to evaluate accuracy based on question types.

**Multiple-Choice Questions Evaluation Method**

For evaluating the accuracy of multiple-choice questions, we used a direct matching method. Specifically, we optimized the string of the model's output for multiple-choice answers and directly compared the optimized identifier with the standard answer to determine consistency. This method is simple and efficient, allowing for quick evaluation of the model's performance on multiple-choice questions.。

**Fill-in-the-Blank Questions Evaluation Method**

For fill-in-the-blank questions, we implemented an evaluation strategy based on fuzzy logic. This method allows for a certain margin of error in the answers, using fuzzy matching techniques to assess the similarity between the model's output and the standard answer. The core of the fuzzy matching algorithm is its ability to identify and quantify the similarity of key information elements between the answer and the standard answer, even in the presence of slight deviations or different expressions.

Our fuzzy matching technique first uses natural language processing to tokenize and semantically analyze the answers, extracting key information and concepts. By calculating the semantic similarity between keywords in the answer and the standard answer, our algorithm can assess the overall relevance of the answer. This step involves complex semantic embeddings and vector space models[2], such as Word2Vec or BERT, to capture deep semantic relationships between words. For different formats of answers, our fuzzy matching technique considers not only the text content but also the format, structure, and logical flow of the answer to achieve multidimensional evaluation. The final judgment standard includes an adaptive threshold setting mechanism that dynamically adjusts the similarity evaluation standard based on the difficulty of the question and the expected answer range.[3]

**Comprehensive Questions Evaluation Method**

For comprehensive questions, due to the significant format differences, we adopted an innovative evaluation approach using a system of expert evaluation and AI-assisted assessment. This dual evaluation mechanism aims to achieve a comprehensive and objective evaluation of the model's output by combining the efficiency of technology with the deep understanding of experts.

Our AI-assisted evaluation system is built on the latest deep learning frameworks, using multilayer neural networks to understand and assess the complexity of answers. These networks can capture subtle patterns and deep structures in the answers for precise evaluation. The core of the system is an advanced natural language understanding module that can parse and comprehend the complexity of natural language. Through techniques such as semantic role labeling and dependency syntax analysis, the system can deeply understand the semantic content and logical structure of the answers. The logical reasoning engine analyzes the logical consistency of the answers, verifying whether the assumptions, reasoning, and conclusions in the answers are reasonable and consistent with known scientific principles and facts.[4]

To enhance evaluation accuracy, we included expert reviews, using random sampling to select 1/10 of the answers for re-evaluation by domain experts. This sampling mechanism ensures the rigor of the evaluation process and the reliability of the results. Only when a question is completely correct is it considered correct.

4.1.2 Multi-Solution Accuracy

In this study, for the evaluation of multi-solution responses, we adopted a detailed and systematic analysis method. This method extends the evaluation of single-solution responses to accommodate the complexity of multi-solution scenarios. The specific steps we implemented are as follows:

First, we categorized the multi-solution responses from the model based on question types, including multiple-choice, fill-in-the-blank, and comprehensive questions. This step ensures the specificity and effectiveness of subsequent evaluations. Then, for each multi-solution response, we decomposed it into multiple single answers and stored them in an ordered list. We used the Re[5] method to identify possible strings and used a greedy pattern[6] to loop through them, ultimately storing each solution separately. This decomposition strategy allows us to evaluate each potential answer individually, providing a more detailed analysis of the model's creativity and comprehensiveness.

Next, we conducted individual analyses of each decomposed single-solution response. This analysis process is similar to the evaluation method for single-solution responses, including assessments of accuracy, logical reasoning, and completeness. We used advanced natural language processing techniques and machine learning algorithms to ensure the accuracy and consistency of the evaluation.

After evaluating each single-solution response, we aggregated these evaluation results to derive the overall evaluation accuracy of the multi-solution response. To avoid errors and misjudgments, if more than 80% of the solutions in a multi-solution response are correct, we consider it correct.

**4.2 Response Time and Its Evaluation Method**

For the response time of the model, we adopted strict evaluation standards. When the server receives a request, the system adds a timestamp in the code to record this moment, representing the start of the request processing flow. When the server completes the request processing and prepares to send the response, the system adds another timestamp to record this moment, representing the end of the request processing flow. By calculating the difference between these two timestamps, we can determine the total time required for the system to process the request. This time includes the entire process from parsing the request, executing business logic, to generating the response. This method allows us to accurately measure the time required for the model to process a single request, thereby evaluating its performance. For greater precision, we used the 95th percentile response time. We sorted all the response times and determined the value at the 95% position, [7]meaning that 5% of the request response times are longer than this value. This value provides a strict benchmark for evaluating and optimizing system performance.

### 4.3 Logical Reasoning and Its Evaluation Method

To more accurately assess the logical reasoning ability of large language models, we adopted a multi-round chain-of-thought prompting method. The core idea of the chain of thought is that by showing the model examples that include the reasoning process, the model can mimic this process when answering, thereby improving the accuracy of the answers. We set up a mechanism for the model to think before answering, allowing it to automatically use the chain of thought to demonstrate its reasoning ability. Based on this, we evaluated the model's incorrect answers and used the CCoT (Chain of Thought)[1] method to guide the model's responses. CCoT limits the output length, prompting the model to focus on the most critical reasoning steps, reducing irrelevant reasoning and redundant content, and improving the relevance and accuracy of the output. To measure its logical reasoning, we first quantified the model's logical reasoning ability by counting the number of times it successfully used the chain of thought[8] to guide its answers. We then calculated the average number of guidance attempts. For questions with an average guidance attempt of three or more, we considered them successfully guided. Finally, we combined the number of guidance attempts and the success rate:

$$Comprehensive\ Score = (Guidance\ Attempts\ /1 \times 0.6) + (Success\ Rate \times 0.4)$$

This method not only optimizes the model's output quality but also provides an effective means of evaluating its logical reasoning.

### 4.4 Guidance and Its Evaluation Method

For guidance evaluation, we provided a method of multi-prompt guidance to observe the model's response efficiency and the percentage improvement in response accuracy. Specifically, we first determined a set of fixed guiding words to standardize the model's responses. Then, we introduced diverse guiding words and retested to evaluate the model's responsiveness to different guiding words, thereby observing the model's performance in terms of guidance. This process helps us gain a deeper understanding of the model's performance changes under different guiding conditions. The guiding words are set as follows:

**Table 10 Classification table of leading words**

| | |
|---|---|
| Guiding Words A (Mathematician) | "You are an expert with profound knowledge in the field of mathematics, covering everything from basic mathematics to cutting-edge mathematical theories. You excel at discovering the beauty of mathematics and solving complex mathematical problems through rigorous logic and innovative methods. You |
| | need to use advanced methods from the mathematical community to solve this problem: <br> 1. There are no restrictions on the solving methods; utilize higher-difficulty solutions as much as possible." |
| Guiding Words B (College Entrance Examination Student) | "You are a student facing the pressure of the college entrance examination, with a strong interest in mathematics, but sometimes encounter difficult problems. You are eager to improve your mathematical abilities. When faced with this problem, you need to enhance efficiency by using the simplest and most correct methods to ensure you have enough time to solve other questions during the exam." |
| Guiding Words C (Mathematics Teacher) | "You are an experienced mathematics teacher, proficient in using the Socratic method of education to help students solve mathematical problems. I will give you <question>, and after solving it, you need to reconstruct the thought process to assist students in their problem-solving approach: <br> 1. Do not respond to topics unrelated to the question. If <student's question> is unrelated to the question, always reply with 'I still do not understand this question.' <br> 2. Carefully analyze <student's question> and provide appropriate guidance. Start with a simple prompt, then gradually delve deeper, offering more specific problem-solving strategies." |

## 4.5 Creativity (Multiple Solutions)

In evaluating the creative outputs of large models, we constructed an advanced multi-parameter evaluation system that encompasses five core dimensions: correctness, solution diversity, solution richness, time taken, and solution complexity. These dimensions collectively form a comprehensive evaluation model aimed at thoroughly quantifying the performance of large models in creative problem-solving.

Correctness, as the foundation of the evaluation, is achieved through a precise matching algorithm that compares the model's output with a predefined set of standard answers to ensure

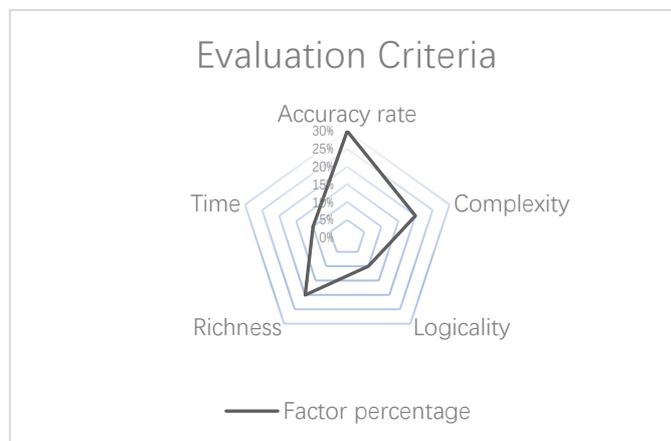

accuracy and reliability.

When assessing solution richness, we employed Regular Expressions (re) technology to semantically segment multi-solution responses, parsing complex multi-solution responses into a set of individual solutions. By calculating the length of the resulting set, we can quantify the model's ability to provide diverse solutions. To encourage the model to generate as many multi-solution responses as possible, we adjusted the model's Temperature to increase randomness.

The evaluation of solution complexity focuses on identifying and measuring conceptual differences between different solutions. This evaluation involves quantifying the depth and structural complexity of solutions. We deployed an advanced AI and expert-assisted evaluation system that can analyze and compare various solutions using deep learning algorithms to identify essential differences between them. Solution diversity is evaluated using a score (1-100 points); a score close to 100 indicates that the response is complex, unique, and diverse, with each answer offering varied approaches to the problem. A score close to 0 suggests high repetition, similar response methods, and simple, chaotic issues.

The evaluation of time taken for solutions focuses on the response time of the model's multi-solution replies. This study implemented a pattern similar to single-solution timing. Whenever the server receives a request, the system embeds a timestamp in the corresponding code segment to capture and record that specific moment, marking the start of the request processing sequence. Once the server completes processing the request and is ready to return a response, the system embeds another timestamp to capture and record that specific moment, marking the end of the request processing sequence. By comparing the difference between these two timestamps, the total time taken by the system to process the request can be accurately calculated. This time encompasses the entire process from parsing the request, executing business logic, to generating the response.

The evaluation of solution logic focuses on the comprehensibility of the model's responses. This evaluation is achieved by combining an AI-assisted evaluation system with insights from domain experts. The system analyzes the structure of solutions through complex algorithms, while experts provide professional assessments of the solutions' innovativeness and practicality. Similar to the logical methods mentioned above, the model's logic is judged by the number of guidance attempts and the success rate of guidance.

## 5 Results Analysis

### 5.1 Time Evaluation

5.1.1 Average Time Taken to Solve Different Question Types in Single-Solution Scenarios

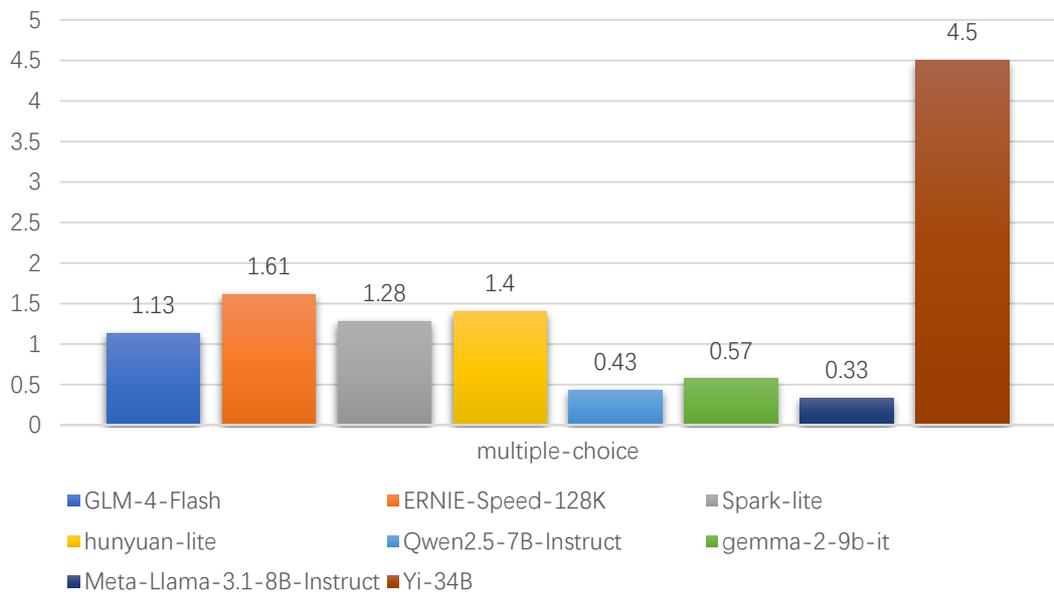

**Fig.1 Average time spent on multiple-choice problem solving**

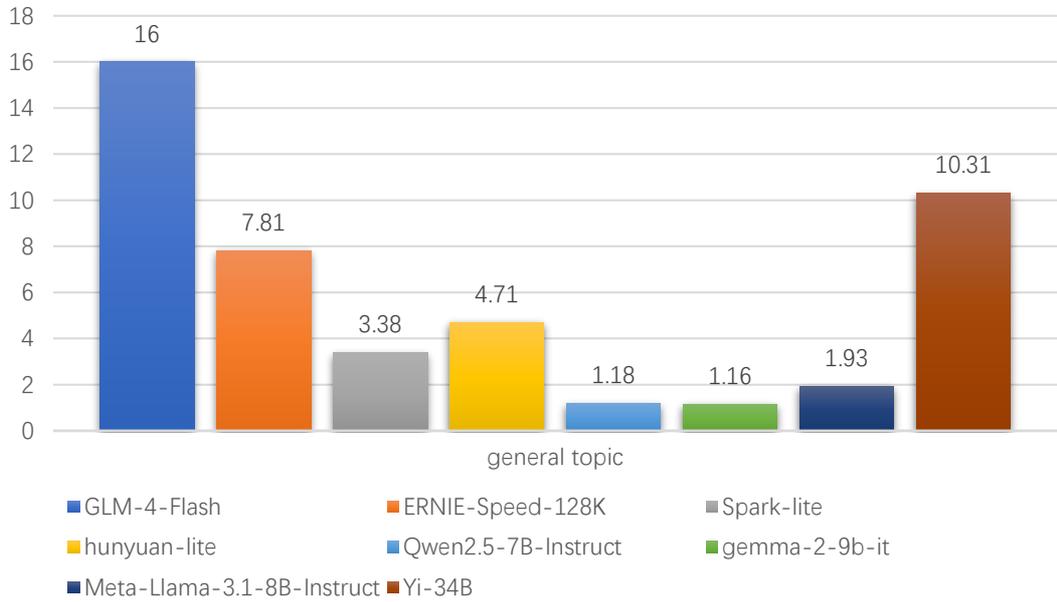

**Fig.2 Average time spent solving fill-in-the-blank questions**

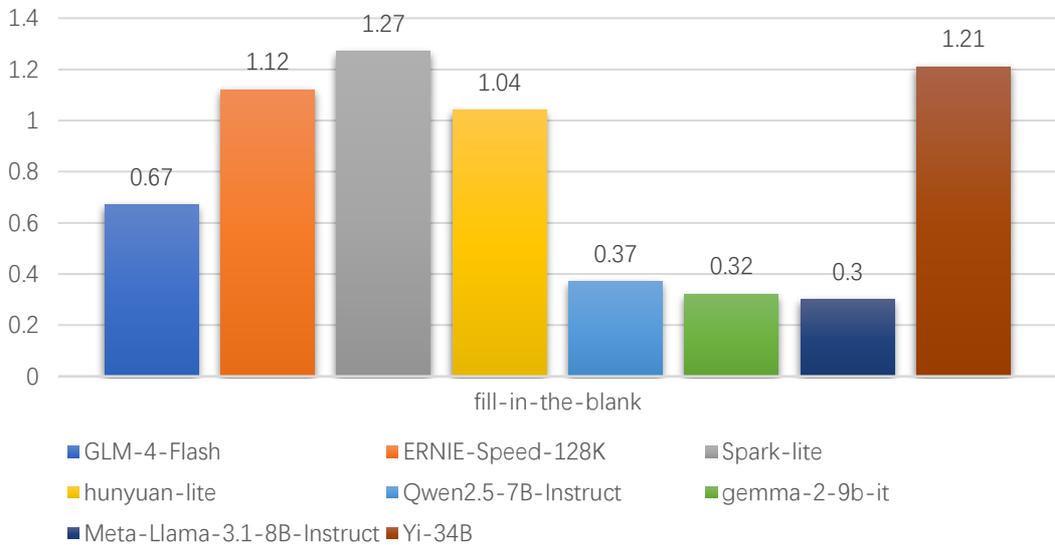

**Fig.3 Average time spent solving general topic questions**

**Table 10 Data synthesis tables**

| Model Name | Multiple-Choice Questions | Fill-in-the-Blank Questions | Question-and-Answer Questions |
|---|---|---|---|
| GLM-4-Flash | 1.12 | 0.67 | 16.01 |
| ERNIE-Speed-128K | 1.61 | 1.12 | 7.81 |
| Spark-lite | 1.28 | 1.27 | 3.38 |
| Hunyuan-lite | 1.40 | 1.04 | 4.71 |
| Qwen2.5-7B-Instruct | 0.43 | 0.36 | 1.18 |
| gemma-2-9b-it | 0.57 | 0.32 | 1.16 |
| Meta-Llama-3.1-8B-Instruct | 0.32 | 0.93 | 1.93 |
| Yi-34B | 4.50 | 1.21 | 10.31 |

From Table 10, we can see that the Qwen2.5-7B-Instruct model performs the best on multiple-choice questions, demonstrating extremely high efficiency. The gemma-2-9b-it model takes the shortest time on fill-in-the-blank questions, while Meta-Llama-3.1-8B-Instruct takes the shortest time on multiple-choice questions. GLM-4-Flash takes the longest time on question-and-answer tasks, whereas Yi-34B takes the longest time on multiple-choice questions.

5.1.2 Average Time Taken to Solve Questions of Different Difficulty Levels in Single-Solution Scenarios

Unit（second /s）

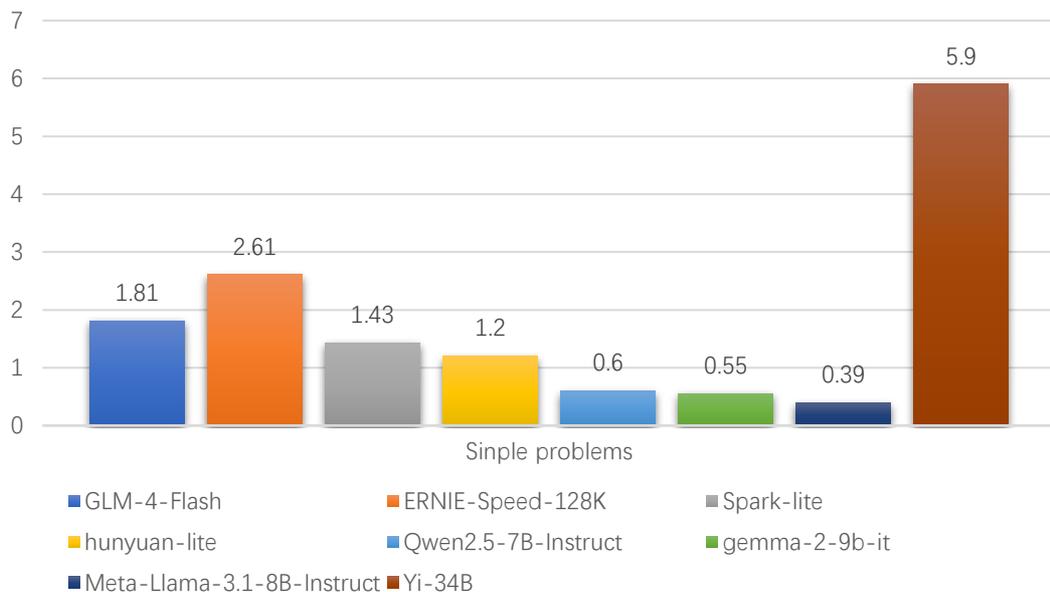

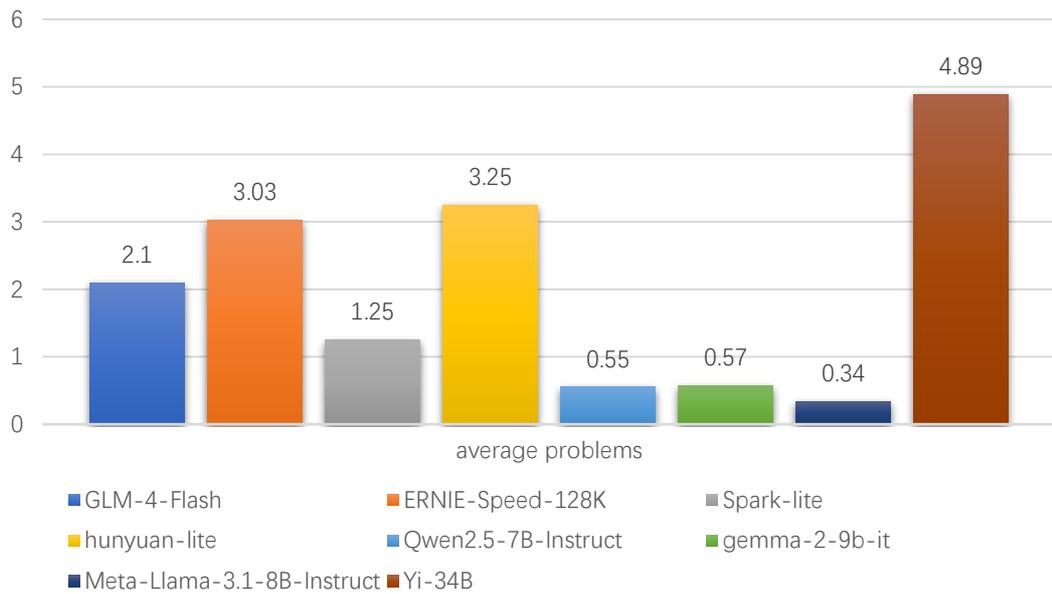

Fig.4 Average time spent solving simple problems

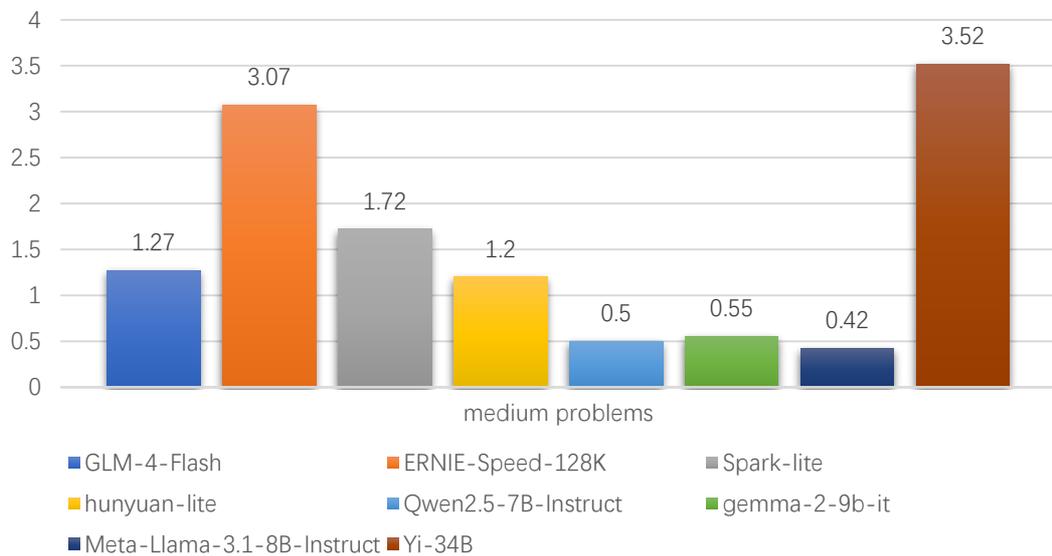

Fig.5 Average time spent solving average problems

Fig.6 Average time spent solving medium problems

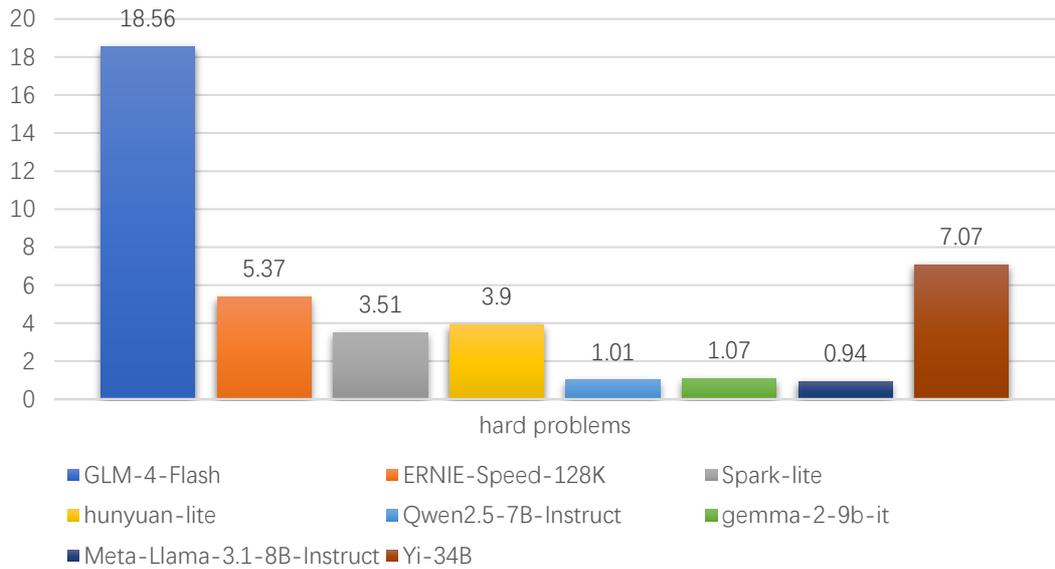

**Fig.7 Average time spent solving hard problems**

**Table 11 Data synthesis tables**

| Model Name | Multiple-Choice Questions | Fill-in-the-Blank Questions | Question-and-Answer Questions | 难题 |
|---|---|---|---|---|
| GLM-4-Flash | 1.81 | 1.27 | 2.10 | 18.56 |
| ERNIE-Speed-128K | 2.61 | 3.07 | 3.03 | 5.37 |
| Spark-lite | 1.43 | 1.72 | 1.25 | 3.51 |
| Hunyuan-lite | 1.20 | 1.20 | 3.25 | 3.90 |
| Qwen2.5-7B-Instruct | 0.60 | 0.50 | 0.55 | 1.01 |
| gemma-2-9b-it | 0.55 | 0.55 | 0.57 | 1.07 |
| Meta-Llama-3.1-8B-Instruct | 0.39 | 0.42 | 0.34 | 0.94 |
| Yi-34B | 5.90 | 3.52 | 4.89 | 7.07 |

From Table 11, we can see that the Qwen2.5-7B-Instruct and gemma-2-9b-it models perform excellently when handling simple and intermediate questions, with the shortest processing times. The Meta-Llama-3.1-8B-Instruct model also performs well across all difficulty levels, particularly showing the shortest processing times for simple and advanced questions. The Yi-34B model has longer processing times across all difficulty levels, especially taking the longest time for simple questions, indicating that it may require more computational resources or time when processing questions.

### 5.1.3 Multi-Solution Time Evaluation

Unit (second /s)

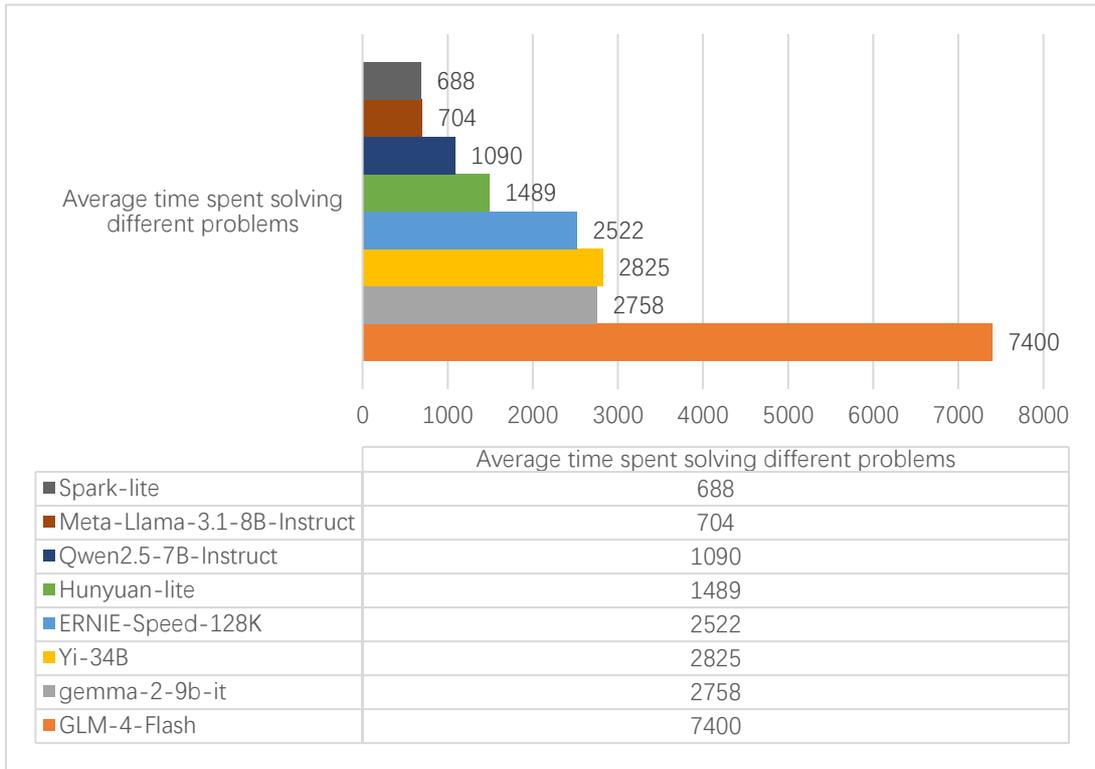

**Fig.8 Average time spent solving different problems**

By comparing Figure 8 and conducting detailed data analysis, the Spark-lite model shows the lowest processing time, at only 688 seconds, while the GLM-4-Flash model has the longest processing time, reaching 7400 seconds. The processing times for other models are as follows: Meta-Llama-3.1-8B-Instruct model 704 seconds, Qwen2.5-7B-Instruct model 1090 seconds, Hunyuan-lite model 1489 seconds, LRNE-Speed-128K model 2522 seconds, Yi-34B model 2825 seconds, and gemma-2-9b-it model 2758 seconds.

From the data, it can be inferred that the Spark-lite model has a significant advantage in efficiency when handling multi-solution problems, while the GLM-4-Flash model has considerable room for improvement in efficiency when dealing with multi-solution problems.

### 5.2 Accuracy Evaluation

### 5.2.1 Evaluation of correct rate of different question types under single solution condition

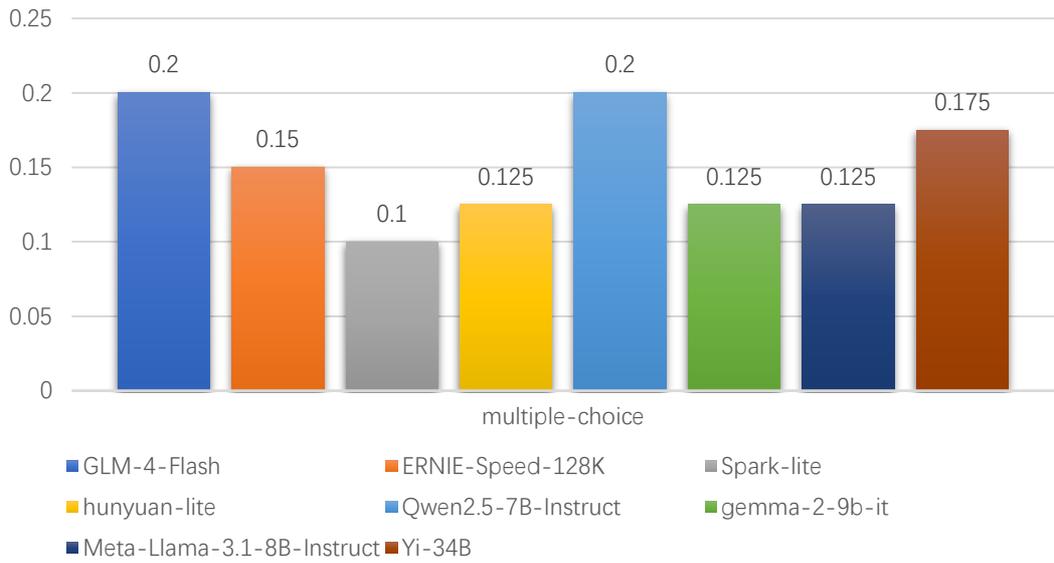

**Fig.9 Average correctness of multiple-choice questions**

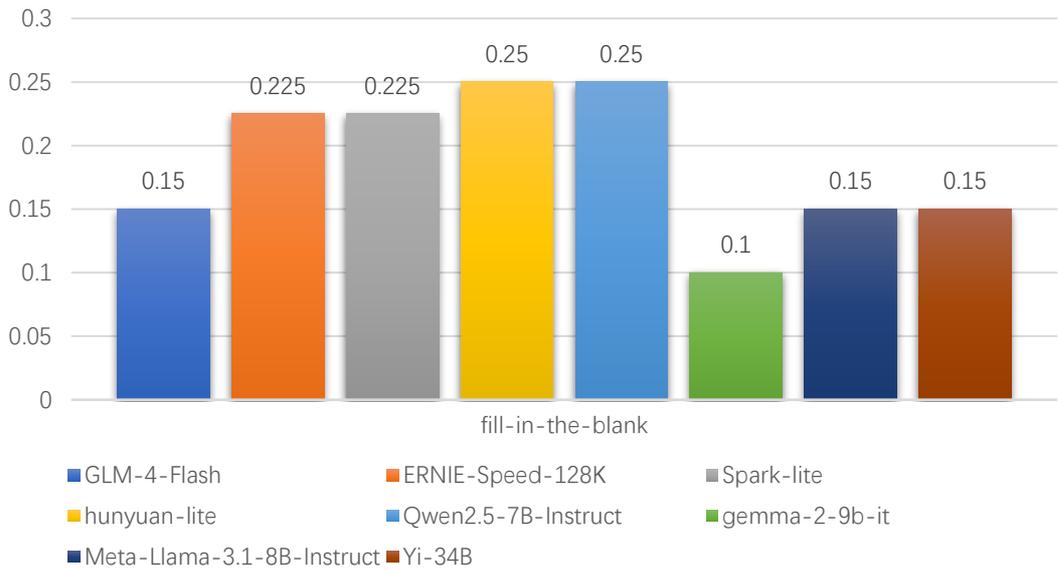

**Fig.10 Average correctness of multiple-choice questions**

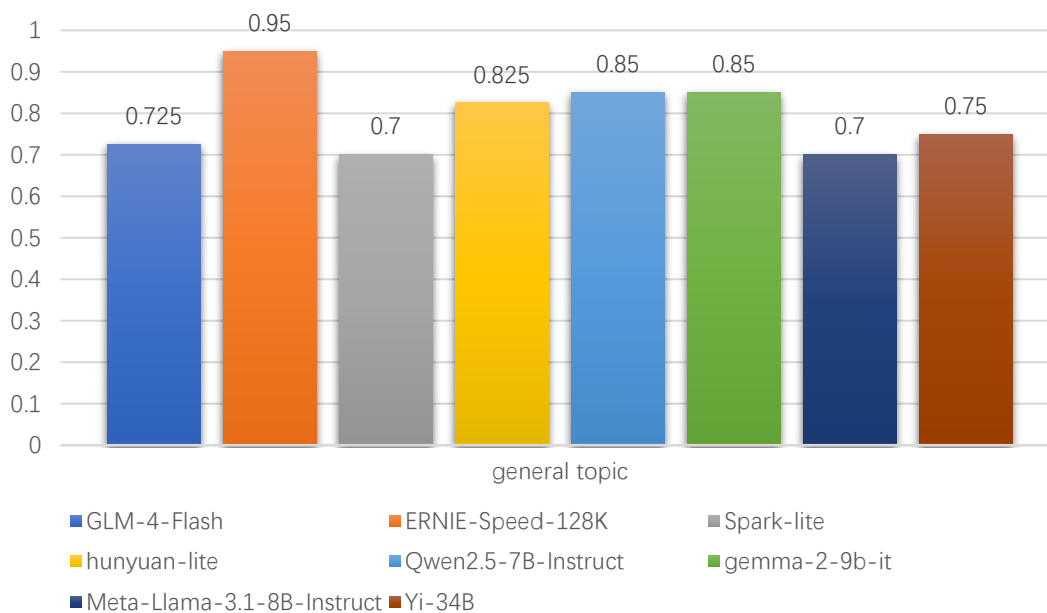

**Fig.11 Average correctness of general topic questions**

**Table 12 Data synthesis tables**

| Model Name | Multiple-Choice Accuracy | Fill-in-the-Blank Accuracy | Comprehensive Question Accuracy |
| --- | --- | --- | --- |
| GLM-4-Flash | 0.200 | 0.150 | 0.725 |
| ERNIE-Speed-128K | 0.150 | 0.225 | 0.950 |
| Spark-lite | 0.100 | 0.225 | 0.700 |
| Hunyuan-lite | 0.125 | 0.250 | 0.825 |
| Qwen2.5-7B-Instruct | 0.200 | 0.250 | 0.850 |
| gemma-2-9b-it | 0.125 | 0.100 | 0.850 |
| Meta-Llama-3.1-8B-Instruct | 0.125 | 0.150 | 0.700 |
| Yi-34B | 0.175 | 0.150 | 0.750 |

From Table 12, it can be seen that the ERNIE-Speed-128K and Qwen2.5-7B-Instruct models perform the best on comprehensive questions, demonstrating high accuracy when handling complex problems. Hunyuan-lite has the highest accuracy on fill-in-the-blank questions, while Spark-lite has the lowest accuracy on multiple-choice questions.

## 5.2.2 Accuracy for Different Difficulty Levels in Single-Solution Scenarios

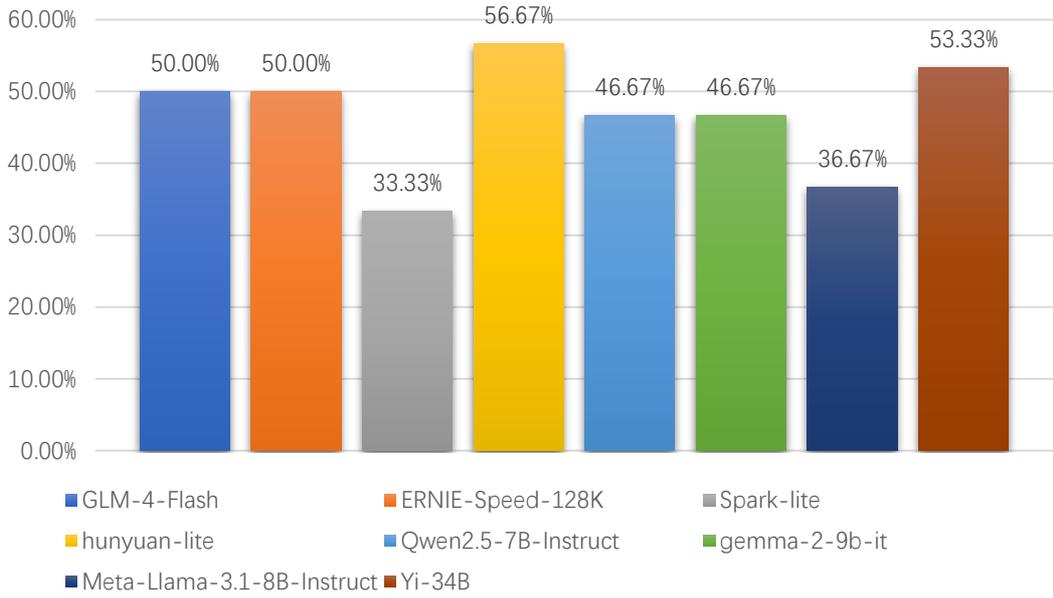

**图 12 简单题平均正确率**

**Fig.12 Average correctness of simple questions**

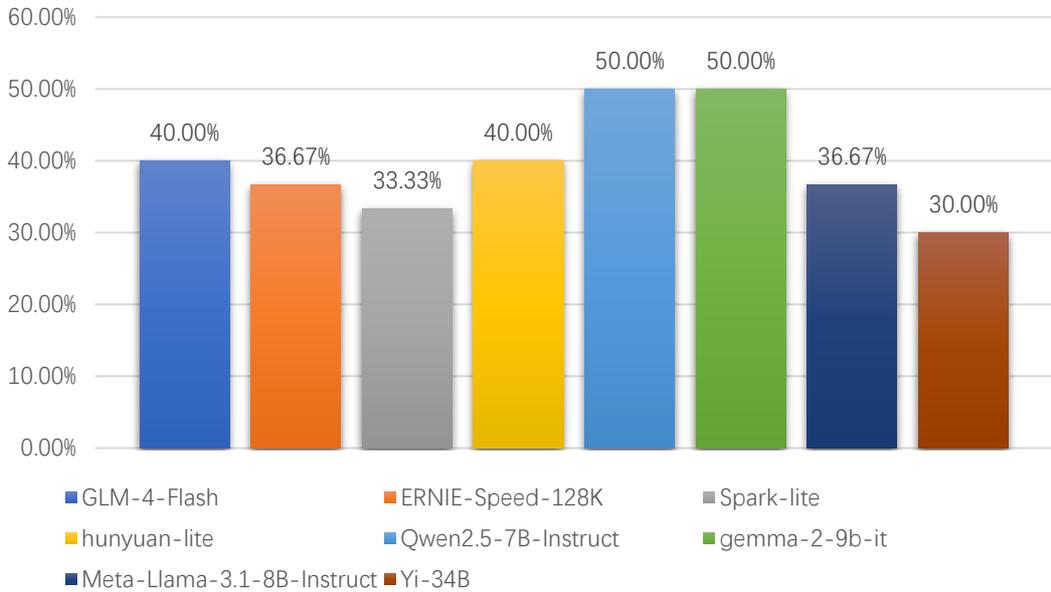

**图 13 中档题平均正确率**

**Fig.13 Average correctness of average questions**

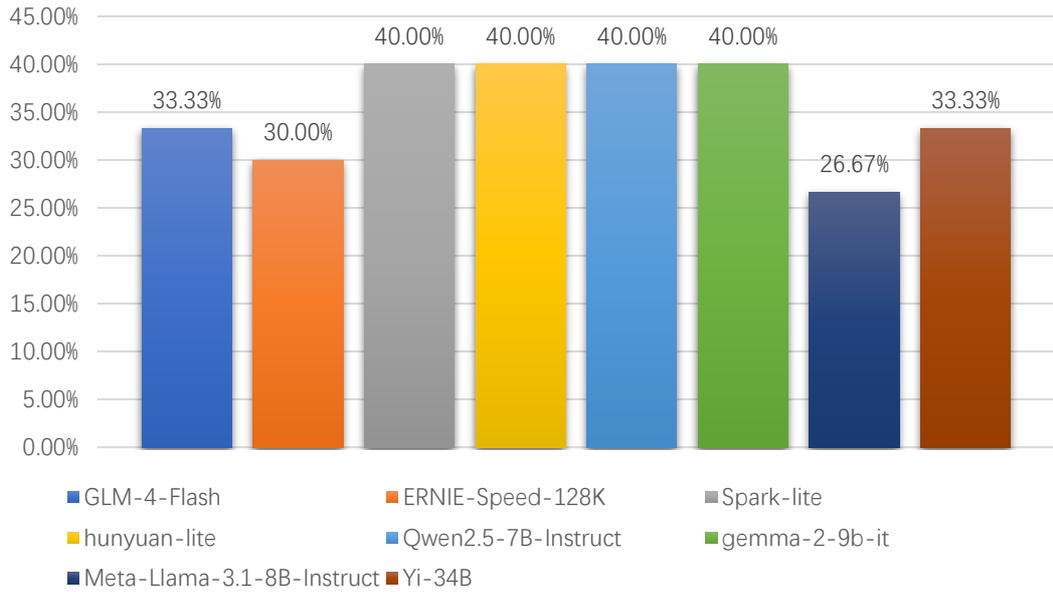

Fig.14 Average orrectness of medium questions

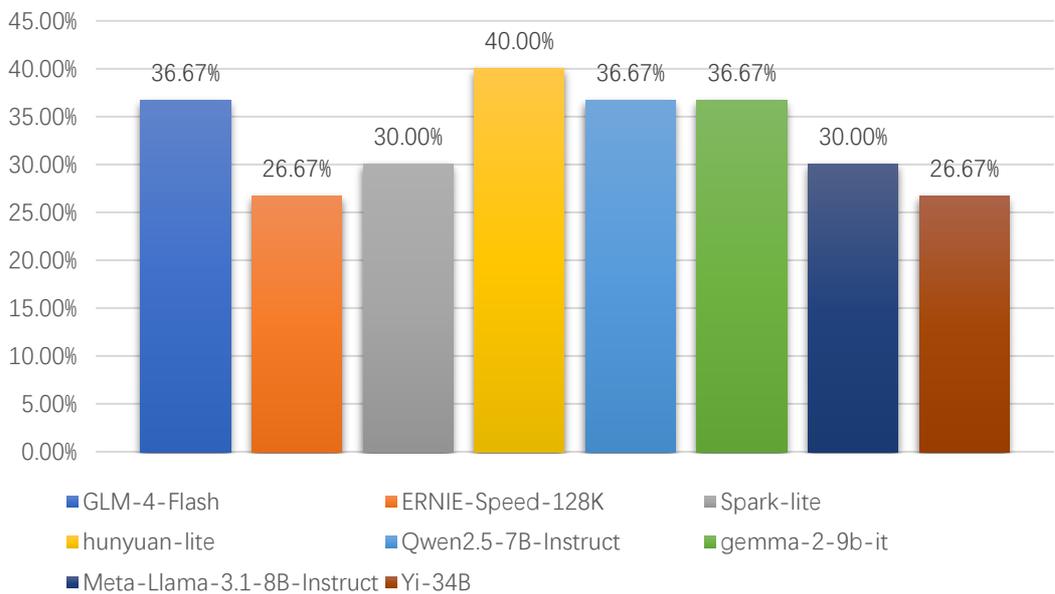

Fig.15 Average orrectness of hard questions

Table 13 Data synthesis tables

| Model Name | Simple Questions | Intermediate Questions | Advanced Questions | Difficult Questions |
|---|---|---|---|---|
| GLM-4-Flash | 50.00% | 40.00% | 33.33% | 36.67% |
| ERNIE-Speed-128K | 50.00% | 36.67% | 30.00% | 26.67% |
| Spark-lite | 33.33% | 33.33% | 40.00% | 30.00% |
| Hunyuan-lite | 56.67% | 40.00% | 40.00% | 40.00% |
| Qwen2.5-7B-Instruct | 46.67% | 50.00% | 40.00% | 36.67% |
| gemma-2-9b-it | 46.67% | 50.00% | 40.00% | 36.67% |
| Meta-Llama-3.1-8B-Instruct | 36.67% | 36.67% | 26.67% | 30.00% |
| Yi-34B | 53.33% | 30.00% | 33.33% | 26.67% |

As shown in Table 13, Hunyuan-lite and Qwen2.5-7B-Instruct excel in simple and intermediate questions, demonstrating high accuracy when handling basic and moderately difficult questions. GLM-4-Flash and ERNIE-Speed-128K perform excellently on simple questions, but their accuracy decreases on advanced and difficult questions. Spark-lite and Meta-Llama-3.1-8B-Instruct show relatively balanced performance across all difficulty levels, though their accuracy on difficult questions is lower. Yi-34B performs best on simple questions, but its performance on advanced and difficult questions needs improvement.

5.2.3 Multi-Solution Accuracy Evaluation

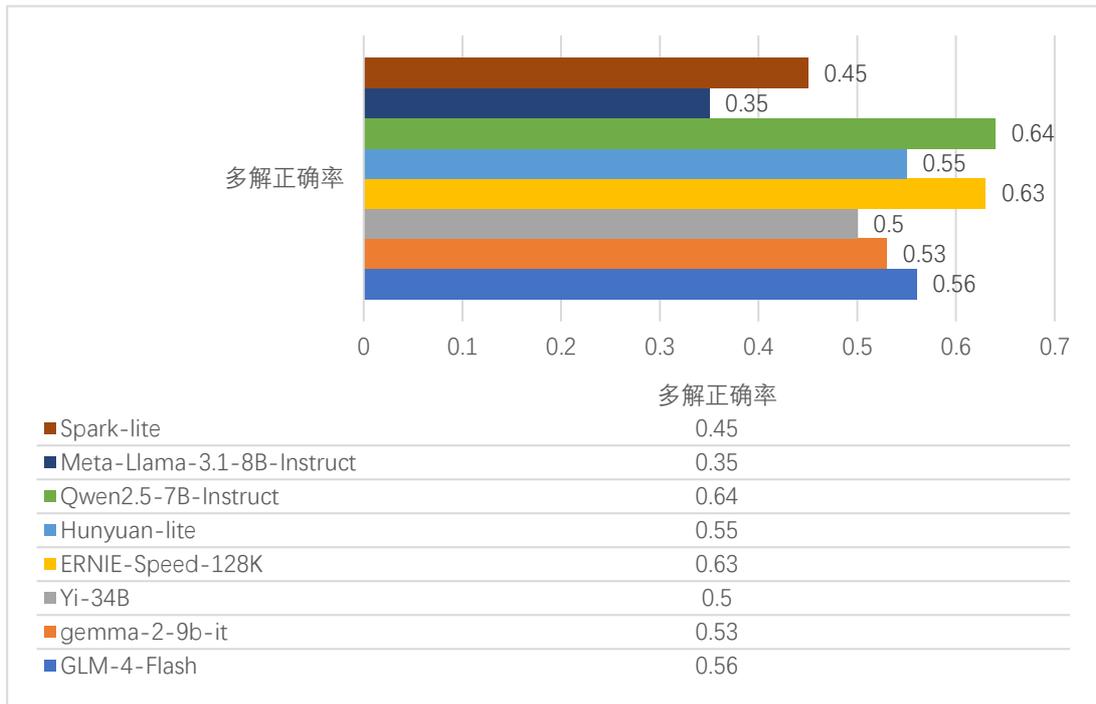

Fig.16 Average orrectness of different questions

As shown in Table 16, the Qwen2.5-7B-Instruct model performs the best in terms of multi-solution accuracy, achieving a correctness rate of 0.64. This means that among all tested multi-solution problems, this model can correctly identify 64% of all possible correct answers. The ERNIE-Speed-128K model follows closely with a correctness rate of 0.63, slightly lower than that of the Qwen2.5-7B-Instruct model.

The correctness rates of the other models range from 0.45 to 0.56. Both Spark-lite and Meta-Llama-3.1-8B-Instruct models have the same correctness rate of 0.45, which is the lowest among all models. Meanwhile, the correctness rates for Hunyuan-lite, Yi-34B, gemma-2-9b-it, and GLM-4-Flash models are 0.55, 0.55, 0.53, and 0.56, respectively, indicating that these models perform relatively similarly on multi-solution problems, but all are lower than those of the Qwen2.5-7B-Instruct and ERNIE-Speed-128K models.

## 5.3 Logic Evaluation

5.3.1 Average Attempts and Success Rate for Guiding Incorrect Questions

Unit （per time）

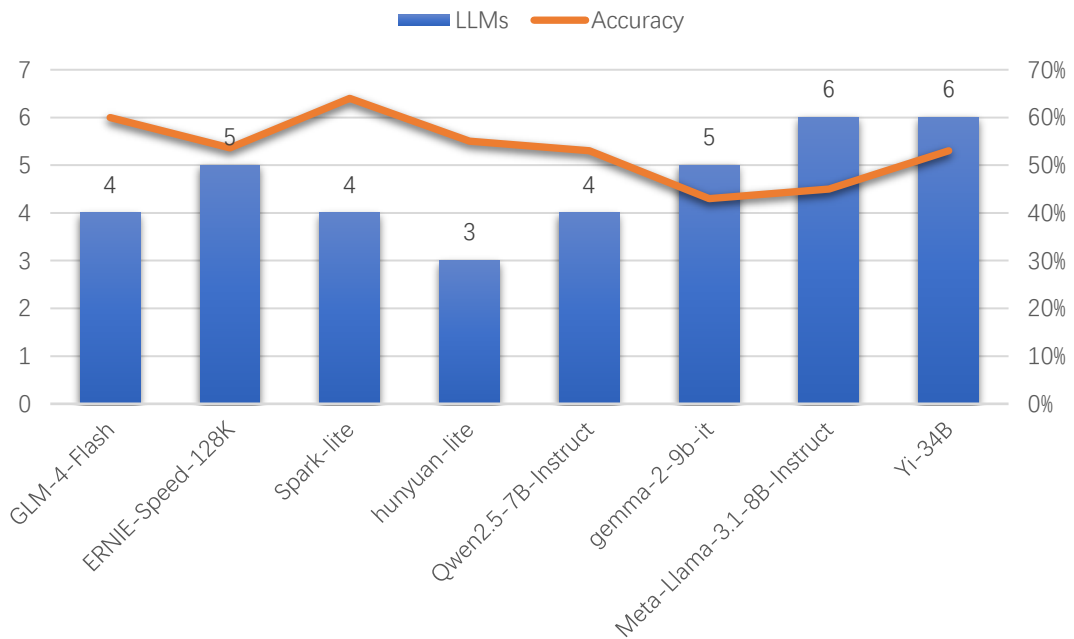

Fig.17 Average number of successful bootstrapping attempts for incorrect topics

From Figure 17, it can be observed that the Hunyuan-lite model performs the best on guiding erroneous questions, requiring an average of only 3 attempts to successfully guide to the correct answer while maintaining a high success rate. Following closely are the Spark-lite, Qwen2.5-7B-Instruct, and gemma-2-9b-it models, all of which require 4 attempts, though their success rates vary slightly. The performances of GLM-4-Flash and ERNIE-Speed-128K are relatively close, requiring 4 and 5 attempts for successful guidance, respectively, with a corresponding decrease in success rates. The Meta-Llama-3.1-8B-Instruct model requires 5 attempts, with a reduced success rate. The Yi-34B model requires the most attempts, at 6, but maintains a relatively high success rate.

5.3.2 Logic Summary

Logic Evaluation Table for Different Models

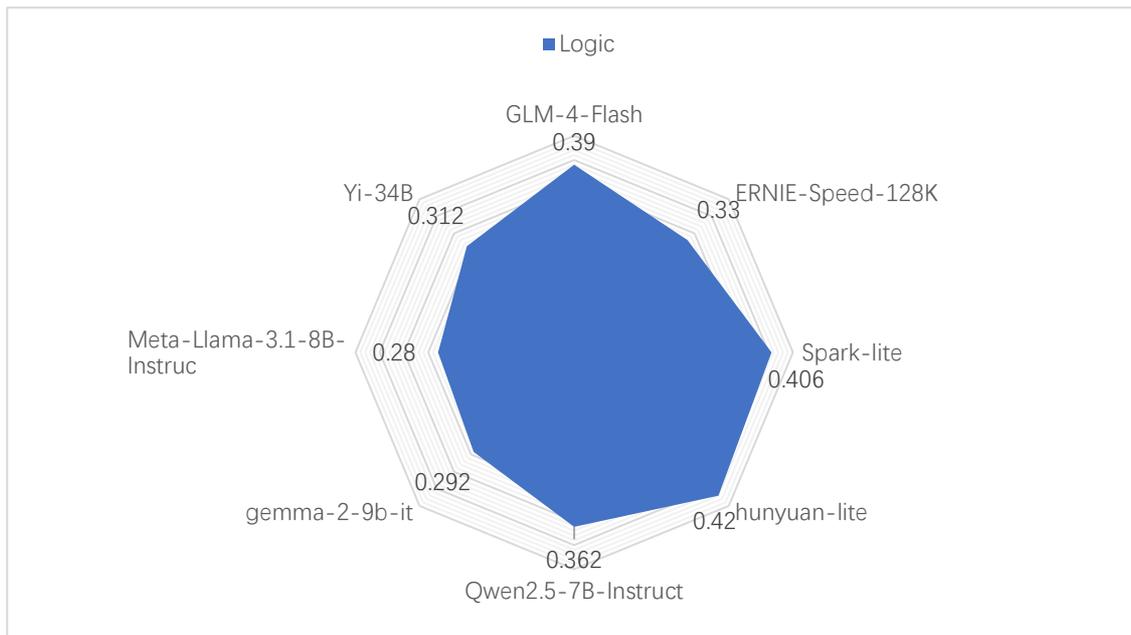

**Fig.18 Large Model Logic Review Chart**

From Figure 18, it can be seen that the Hunyuan-lite model performs the best, indicating that it has the highest efficiency in successfully guiding erroneous questions with the fewest attempts. Following closely are the Spark-lite and GLM-4-Flash models, which also demonstrate high efficiency and success rates. These results suggest that when guiding erroneous questions, the Hunyuan-lite model not only requires fewer attempts but also achieves a high success rate, which may indicate its superior understanding and guiding ability when dealing with logical problems. Other models, such as Spark-lite and GLM-4-Flash, also perform well, but their efficiency is slightly lower compared to Hunyuan-lite.

### 5.4 Guided Evaluation

5.4.1 Evaluation Report for Prompt A (Mathematician) and Prompt C (Math Teacher

As shown in Figure 19, the use of different prompts has led to improvements in accuracy for most models, with Spark-lite showing the most significant increase of 13%. Both Hunyuan-lite and Qwen2.5-7B-Instruct also demonstrated a 12% improvement. In terms of speed enhancement, the performance of most models varied; some models, such as Meta-Llama-3.1-8B-Instruct, Yi-34B, and GLM-4-Flash, experienced a decrease in speed, with reductions of 10% and 12%, respectively. Conversely, ERNIE-Speed-128K improved its speed by 5%, indicating that this model optimized processing speed while maintaining or increasing accuracy, thus achieving greater efficiency.

In summary, Spark-lite performed the best in terms of accuracy improvement, while Hunyuan-lite and Qwen2.5-7B-Instruct also showed excellent results. However, Meta-Llama-3.1-8B-Instruct and Yi-34B experienced a decline in speed enhancement, which may suggest that they sacrificed some processing speed in favor of improving accuracy. This indicates that under the conditions of prompts from mathematicians and math teachers, the models tend to engage in more prolonged and conservative problem-solving thought processes, yet still achieve a notable improvement in answer accuracy.

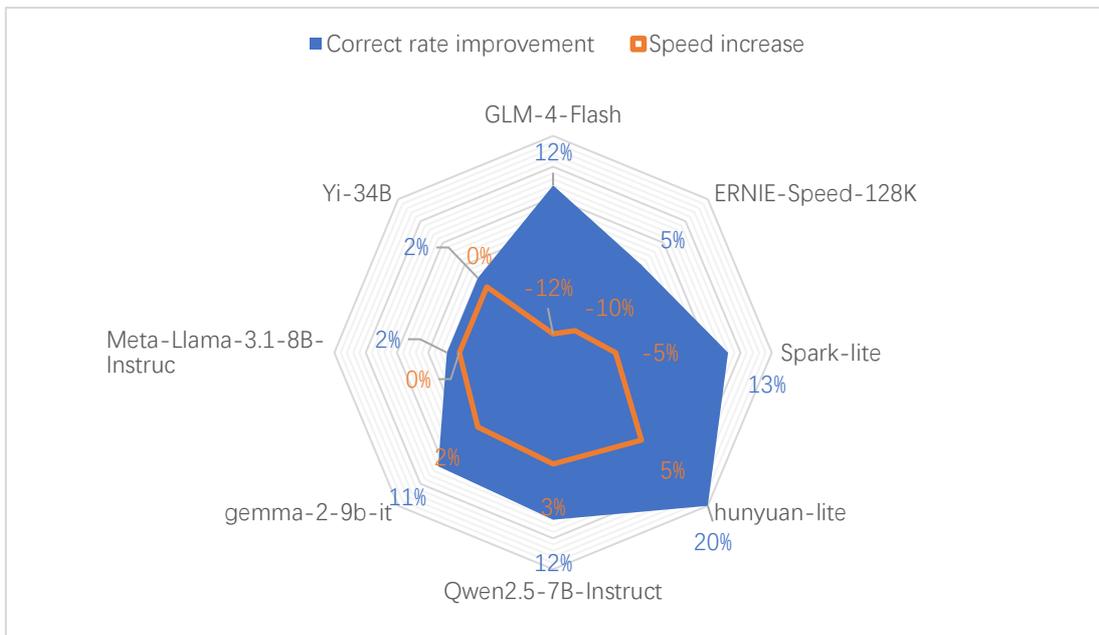

**Fig.19 Guiding Words (A,C) Guided Review Chart**

### 5.4.2 Guiding Words B (College Entrance Examination Students) Evaluation Report

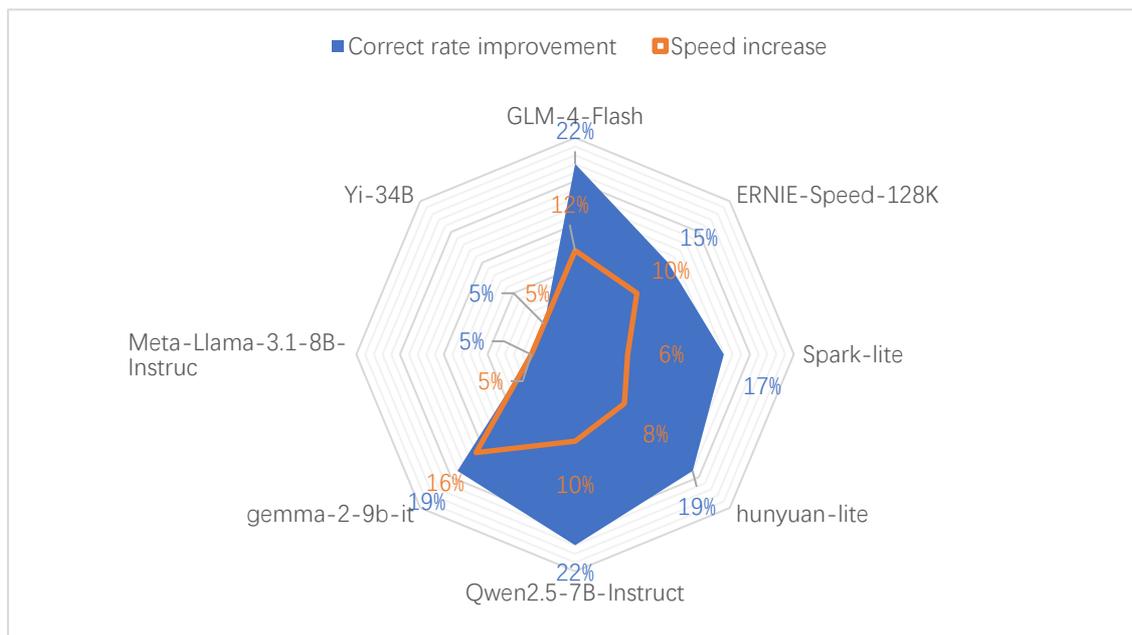

**Fig.20 Guiding Words (B) Guided Review Chart**

As shown in Figure 20, under the guidance of Guiding Words B, most models exhibit a significant improvement in accuracy. Among them, the GLM-4-Flash model shows the most notable increase, reaching 22%. The ERNIE-Speed-128K and Spark-lite models also demonstrate high accuracy improvements of 15% and 17%, respectively. Other models, such as Qwen2.5-7B-Instruct, Hunyuan-lite, gemma-2-9b-it, and Meta-Llama-3.1-8B-Instruct, show accuracy improvements of 10%, 8%, 16%, and 5%, respectively.

In terms of speed improvement, the performance of the models varies significantly. Yi-34B and gemma-2-9b-it show the most notable speed improvements, at 19% and 16%, respectively. The ERNIE-Speed-128K and Spark-lite models also show significant speed improvements of 10% and 6%. In contrast, Qwen2.5-7B-Instruct and Hunyuan-lite have relatively small speed improvements of 5% and 8%, respectively. The Meta-Llama-3.1-8B-Instruct model shows a speed improvement of 5%.

In the evaluation of Guiding Words B for the college entrance examination student group, GLM-4-Flash performs best in terms of accuracy improvement, while ERNIE-Speed-128K and Spark-lite also perform well. In terms of speed improvement, Yi-34B and gemma-2-9b-it stand out. This indicates that under the guidance of college entrance examination students, the models are better able to grasp the key points of the questions, resulting in higher efficiency and accuracy in problem-solving.

### 5.5 Creative Evaluation

## 5.5.1 Richness Evaluation of Different Large Models in Multi-Solution Scenarios

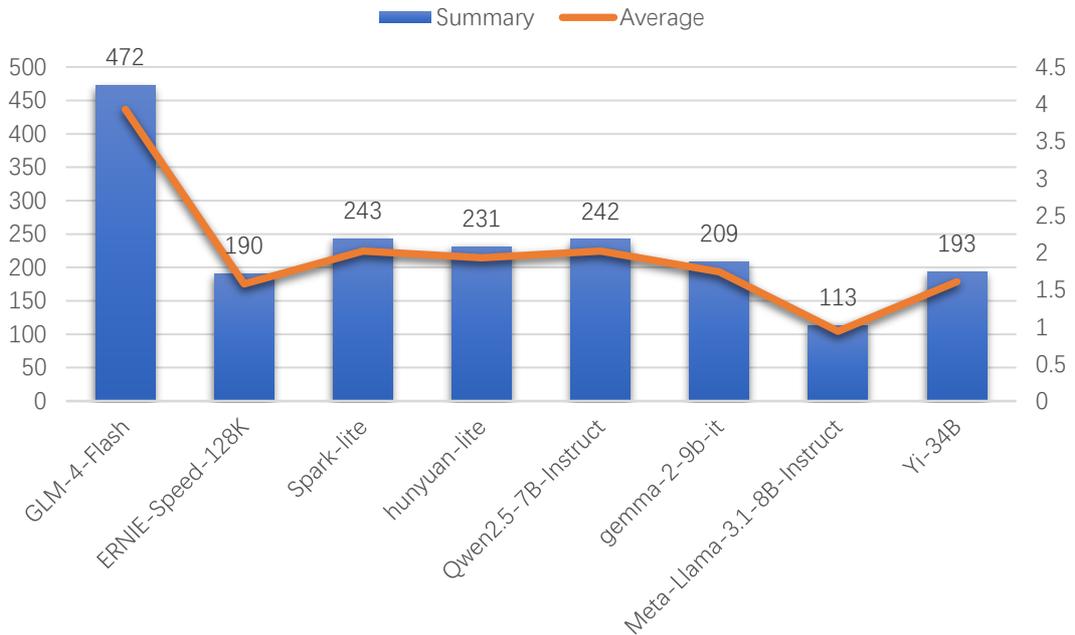

Fig.21 Measurement of the number of solutions in the multiple-solution case for different large models

From Figure 21, it can be seen that the GLM-4-Flash model stands out in total number of solutions, reaching 472 questions. The ERNIE-Speed-128K model follows closely with a total of 190 questions. The total number of solutions for other models is as follows: Spark-lite (243 questions), Hunyuan-lite (231 questions), Qwen2.5-7B-Instruct (242 questions), gemma-2-9b-it (209 questions), Meta-Llama-3.1-8B-Instruct (113 questions), and Yi-34B (193 questions).

In terms of averages, GLM-4-Flash also leads with an average of 4.5 solutions per question. The average for ERNIE-Speed-128K is 3.5 solutions, while the averages for other models range from 2.5 to 3.5 solutions.

In summary, in multi-solution scenarios, the GLM-4-Flash model performs best in both total number and average number of solutions, demonstrating its superior performance in handling such problems. The ERNIE-Speed-128K model also performs well, but there is still a gap compared to GLM-4-Flash. The performance of other models is relatively balanced, with minor differences。

### 5.5.2 Complexity Evaluation of Different Large Models in Multi-Solution Scenarios

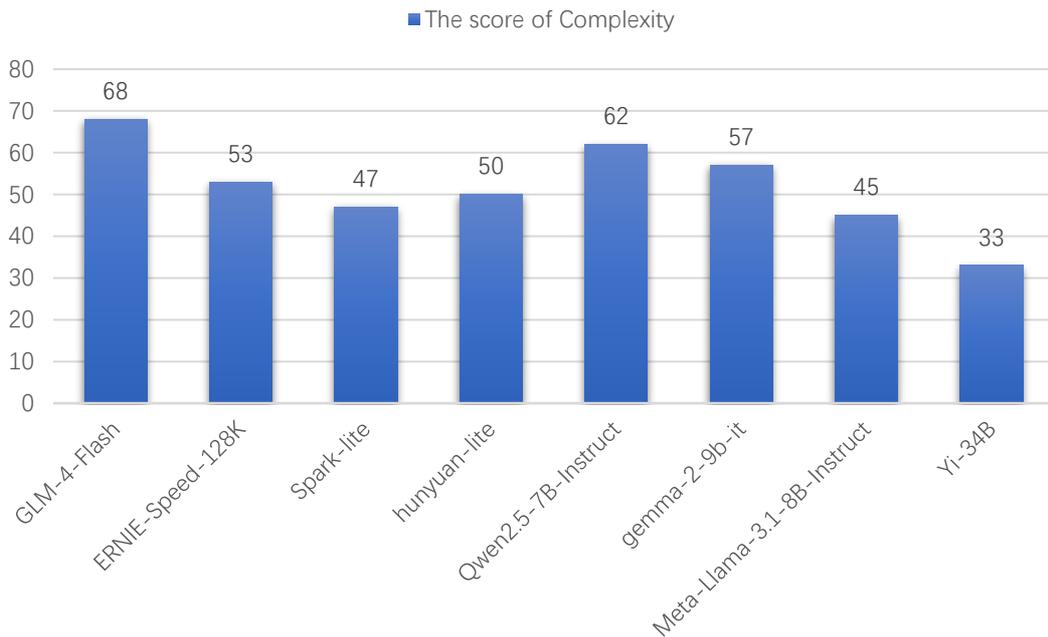

**Fig.22 Complexity Measurement of Different Large Models with Multiple Solutions**

From Figure 22, it can be seen that the GLM-4-Flash model has the highest complexity score of 68, indicating that it exhibits high complexity when handling multi-solution problems. Following closely is the gemma-2-9b-it model with a score of 62. The complexity scores for ERNIE-Speed-128K and Meta-Llama-3.1-8B-Instruct are 53 and 45, respectively, placing them at a moderate level. Spark-lite, Hunyuan-lite, and Qwen2.5-7B-Instruct have scores of 47, 50, and 50, respectively, indicating relatively low complexity. The Yi-34B model has the lowest complexity score of 33, suggesting minimal complexity in handling multi-solution problems.

In handling multi-solution problems, different models exhibit varying levels of complexity. The GLM-4-Flash and gemma-2-9b-it models lead in complexity, indicating that they possess uniqueness in different solutions, with each solution employing different methods. The low complexity score of the Yi-34B model may suggest high redundancy in responses, lacking representativeness in diverse solutions.

### 5.5.3 Comprehensive Evaluation of Diversity Among Different Large Models

Table 14 Diversity Measurement Chart

| Model Name | Time(s) | Accuracy(%) | Complexity | logic | Richness |
|---|---|---|---|---|---|
| GLM-4-Flash | 7400 | 0.56 | 68 | 0.39 | 3.2 |
| ERNIE-Speed-128K | 2522 | 0.63 | 53 | 0.33 | 2.2 |
| Spark-lite | 688 | 0.45 | 47 | 0.406 | 2.5 |
| Meta-Llama-3.1-8B-Instruct | 704 | 0.35 | 45 | 0.28 | 1.3 |
| Qwen2.5-7B-Instruct | 1090 | 0.64 | 62 | 0.362 | 2.4 |
| hunyuan-lite | 1489 | 0.55 | 50 | 0.42 | 2.4 |
| gemma-2-9b-it | 2758 | 0.53 | 57 | 0.292 | 2.2 |
| Yi-34B | 2825 | 0.5 | 33 | 0.312 | 1.8 |

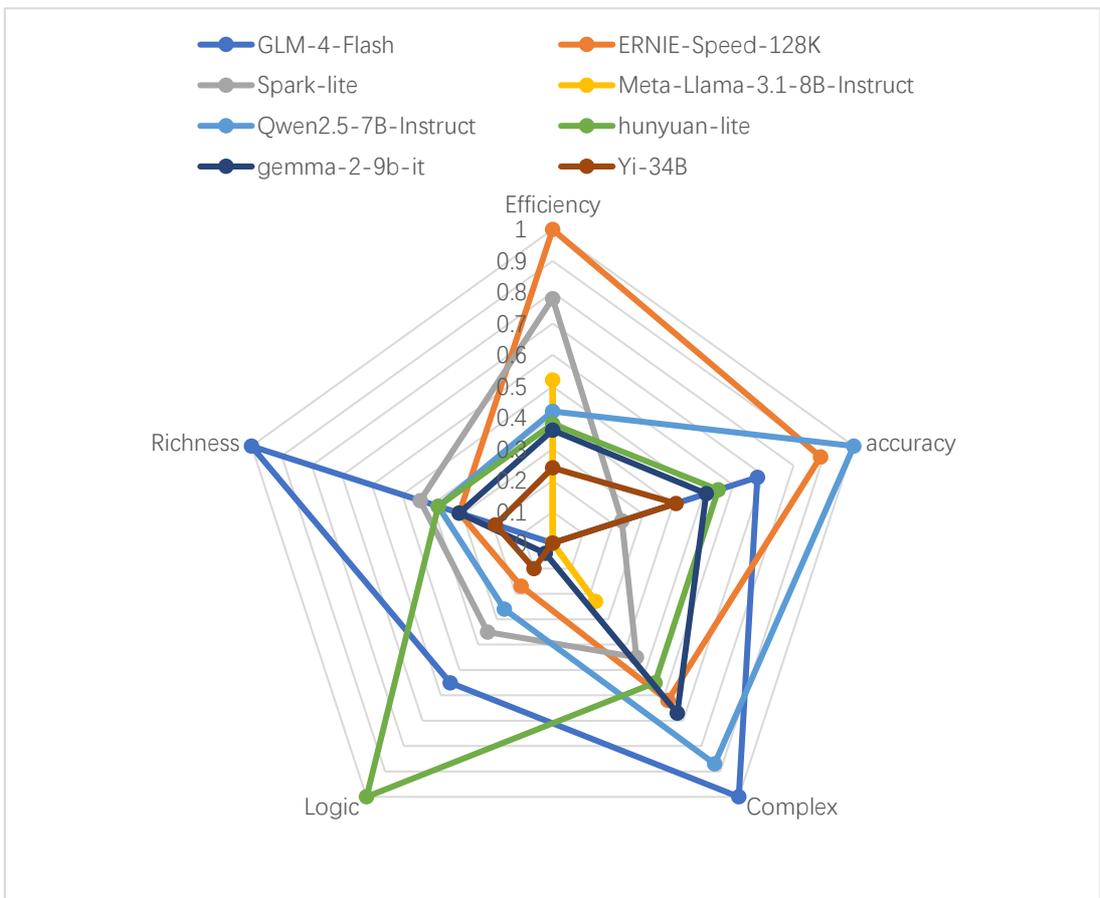

Fig.23 Comprehensive Measurement of Diversity in Different Large Models

From Figure 23, it can be observed that there are significant performance differences among the various models when handling tasks. The ERNIE-Speed-128K and Qwen2.5-7B-Instruct models excel in accuracy, while Spark-lite shows a clear advantage in time efficiency. Although

GLM-4-Flash has a longer processing time, it performs well in terms of accuracy and complexity. The Meta-Llama-3.1-8B-Instruct and Yi-34B models require further optimization across multiple metrics.

For each large model, the comprehensive scores are as follows:

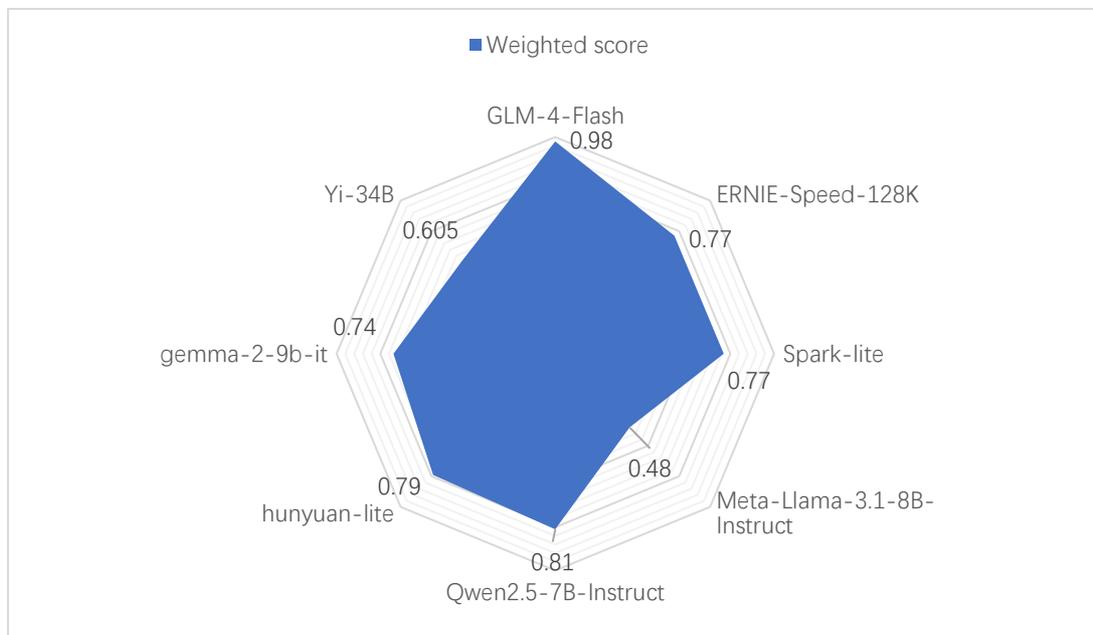

**Fig.24 Creativity comprehensive weighted score**

**Table 15 Large Model Measurement Table**

| | |
|---|---|
| Qwen2.5-7B-Instruct | The model excels in accuracy, complexity, and richness, with a lower logic score but shorter processing time, resulting in a high efficiency score and an overall weighted score of 0.8124917, ranking first. |
| GLM-4-Flash | The model performs well in accuracy and complexity, but due to longer processing time, its efficiency score is lower, leading to an overall weighted score of 0.983135, ranking second. |
| hunyuan-lite | The model shows good performance in accuracy and richness, with a high logic score and moderate processing time, resulting in an overall weighted score of 0.7870671, ranking third. |
| Spark-lite | The model performs average in accuracy and complexity, with a high logic score and short processing time, resulting in a high efficiency score and an overall weighted score of 0.769745, ranking fourth. |
| ERNIE-Speed-128K | The model performs well in accuracy and complexity, but has a low logic score and shorter processing time, resulting in a high efficiency score and an overall weighted score of 0.768396, ranking fifth. |
| gemma-2-9b-it | The model performs average in accuracy and complexity, with a low logic score and longer processing time, resulting in a low efficiency score and an overall weighted score of 0.7422361, ranking sixth. |
| Meta-Llama-3.1-8B-Instruct | The model performs poorly across all metrics, especially in accuracy and logic, with longer |
|   | processing time, resulting in an overall weighted score of 0.483141, ranking seventh. |
| Yi-34B | The model performs average in accuracy and richness, with a low logic score and the longest processing time, resulting in the lowest efficiency score and an overall weighted score of 0.6072355, ranking eighth. |

In summary, the Qwen2.5-7B-Instruct model performs the best when considering all metrics and weights, particularly in terms of accuracy and efficiency. The GLM-4-Flash and Hunyuan-lite models also perform well, ranking second and third, respectively. The Meta-Llama-3.1-8B-Instruct and Yi-34B models underperform across multiple metrics and require further optimization.

### 5.6 Comprehensive Evaluation

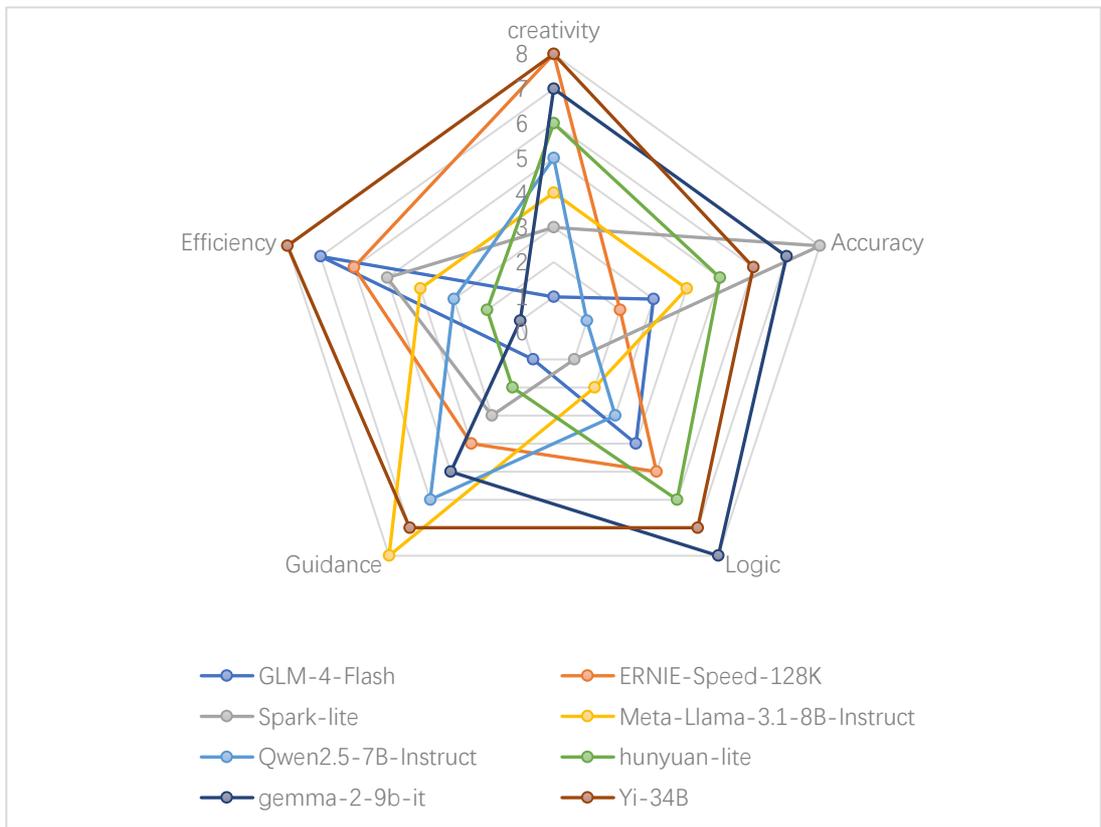

**Fig.25 Creativity comprehensive weighted score**

**Table 16 Large Model Comprehensive Assessment Form**

| Model Name | Creativity | Accuracy | Logic | Guidance | Efficiency |
|---|---|---|---|---|---|
| GLM-4-Flash | 1 | 3 | 4 | 1 | 7 |
| ERNIE-Speed-128K | 8 | 2 | 5 | 4 | 6 |
| Spark-lite | 3 | 8 | 1 | 3 | 5 |
| Meta-Llama-3.1-8B-Instruct | 4 | 4 | 2 | 8 | 4 |
| Qwen2.5-7B-Instruct | 5 | 1 | 3 | 6 | 3 |
| hunyuan-lite | 6 | 5 | 6 | 2 | 2 |
| gemma-2-9b-it | 7 | 7 | 8 | 5 | 1 |
| Yi-34B | 8 | 6 | 7 | 7 | 8 |

**Table 17 Large Model Evaluation Conclusion Table**

| Model | Conclusion |
|---|---|
| GLM-4-Flash | exhibits outstanding performance in both creativity and guidance metrics, particularly ranking first in guidance. This indicates that GLM-4-Flash places significant emphasis on guidance, suggesting that users should focus on diversifying and detailing their prompts to enhance questioning efficiency effectively. |
| ERNIE-Speed-128K | demonstrates strong capabilities in providing accurate answers and has good guidance potential, but its ability to generate novel and innovative answers is relatively weak. |
| Spark-lite | performs best in both accuracy and logic metrics, showcasing its advantage in generating precise and logically sound outputs. This indicates that Spark-lite can not only provide accurate answers but also maintain a rigorous reasoning process when handling tasks. |
| Meta-Llama-3.1-8B-Instruct | It shows strong logical reasoning capabilities and excels in mathematical logic processing; however, its speed in problem-solving is at a moderate level, and it has weaker guidance potential. |
| Qwen2.5-7B-Instruct | ranks first in accuracy alongside Spark-lite but is third in logic. This suggests that while Qwen2.5-7B-Instruct excels in providing accurate answers, it may require further optimization in terms of logical reasoning. |
| hunyuan-lite | ranks second in guidance, demonstrating strong guidance capabilities and high efficiency in handling mathematical problems, indicating its quick problem-solving speed. |
| gemma-2-9b-it | ranks first in efficiency, highlighting its effectiveness in processing mathematical problems. This suggests that gemma-2-9b-it may be more effective in solving mathematical problems that require rapid responses. |
| Yi-34B | ranks low across all metrics, indicating that its performance in |

handling mathematical problems needs improvement. This may suggest that Yi-34B requires further optimization and enhancement in its mathematical problem-solving capabilities.

## 6. Evaluation Summary

This report provides an in-depth exploration of the performance of large language models (LLMs) in solving high school mathematics problems, covering key performance indicators such as accuracy, response time, logical reasoning, guidance, and creativity. These data provide a solid quantitative foundation for assessing the application potential of LLMs in the educational field. Through comparative analysis of the performance of different LLMs, this report reveals their strengths and weaknesses in handling various question types and different difficulty levels, which is crucial for understanding the effectiveness of these models in specific educational contexts. The report not only provides a basis for selecting suitable educational models but also offers guidance for the direction of model training.

The report emphasizes that LLMs have potential for improvement in logical reasoning and recommends enhancing their logical reasoning capabilities through strengthened training and algorithm optimization. The performance of LLMs in creative problem-solving varies, suggesting that expanding diverse training datasets and employing more advanced generative techniques could enhance the models' creativity. Based on the performance data of LLMs, personalized learning pathways can be developed in the future to provide customized learning resources and exercises for students of varying ability levels.

Additionally, this report plans to explore the application of LLMs in other subjects, such as biology and physics, to assess their broader applicability in the educational field. Continuous research is encouraged to track the latest developments in LLM technology and evaluate its long-term impact on educational practices.

## References

1 Nayab, S., Rossolini, G., Buttazzo, G., Manes, N., Giacomelli, F., & Fabrizio, G. (2024). Concise Thoughts: Impact of Output Length on LLM Reasoning and Cost. arXiv:2407.19825.

2 Mikolov, T., Chen, K., Corrado, G., & Dean, J. (2013). Efficient Estimation of Word Representations in Vector Space. In Proceedings of Workshop at ICLR.

3 Liu, Y., & Zhang, Y. (2018). Multi-Dimensional Text Matching with Natural Language Processing Techniques. In Proceedings of SIGIR.

4 McDonald, R., Pereira, F., Ribarov, K., & Hajic, J. (2005). Non-projective dependency parsing using spanning tree algorithms. In Proceedings of HLT/EMNLP.

5 Friedl, J. E. F. (2006). Mastering Regular Expressions. O'Reilly Media.

6 Cormen, T. H., Leiserson, C. E., Rivest, R. L., & Stein, C. (2009). Introduction to Algorithms. MIT Press.

7 Menasce, D. A., & Almeida, V. (2001). Capacity Planning for Web Services: Metrics, Models, and Methods. Prentice Hall.

8 Feng K ,Luo L ,Xia Y , et al.Optimizing Microservice Deployment in Edge Computing with Large Language Models: Integrating Retrieval Augmented Generation and Chain of Thought Techniques[J].Symmetry,2024,16(11):1470-1470.